\documentclass[wcp]{jmlr}

\usepackage{longtable}

\usepackage{booktabs}

\usepackage{lineno}

\usepackage{lipsum}
\usepackage{algorithm}
\usepackage{algorithmic}
\usepackage{multirow}
\usepackage{caption}
\usepackage{wrapfig}

\newcommand{\BibTeX}{B\kern-.05em{\sc i\kern-.025em b}\kern-.08em\TeX}

\makeatletter
\let\Ginclude@graphics\@org@Ginclude@graphics 
\makeatother

\jmlrvolume{}
\jmlryear{2025}
\jmlrworkshop{ACML 2025}

\makeatletter
\def\ps@jmlrtps{%
  \let\@mkboth\@gobbletwo
  \def\@oddhead{}%
  \let\@evenhead\@oddhead
  \def\@oddfoot{}%
  \let\@evenfoot\@oddfoot
}
\makeatother

\title[Activation Steering Meets Preference Optimization]{Activation Steering Meets Preference Optimization: Defense Against Jailbreaks in Vision Language Models}

 \author{\Name{Sihao Wu}$^1$ \Email{sihaowu@liverpool.ac.uk} \\
  \Name{Gaojie Jin}$^2$ \Email{g.jin@exeter.ac.uk}\\
  \Name{Wei Huang}$^3$ \Email{havelhuang@gmail.com} \\
  \Name{Jianhong Wang}$^4$ \Email{jianhong.wang@bristol.ac.uk} \\
  \Name{Xiaowei Huang}$^1$ \Email{xiaoweihuang@liverpool.ac.uk} \\
  \addr $^1$University of Liverpool, $^2$University of Exeter, \\ $^3$Purple Mountain Laboratories,  $^4$University of Bristol.}

\begin{document}

\maketitle

\begin{abstract}
Vision Language Models (VLMs) have demonstrated impressive capabilities in integrating visual and textual information for understanding and reasoning, but remain highly vulnerable to adversarial attacks.
While activation steering has emerged as a promising defence, existing approaches often rely on task-specific contrastive prompts to extract harmful directions, which exhibit suboptimal performance and can degrade visual grounding performance.
To address these limitations, we propose \textit{Sequence-Level Preference Optimization} for VLM (\textit{SPO-VLM}), a novel two-stage defense framework that combines activation-level intervention with policy-level optimization to enhance model robustness. In \textit{Stage I}, we compute adaptive layer-specific steering vectors from diverse data sources, enabling generalized suppression of harmful behaviors during inference. In \textit{Stage II}, we refine these steering vectors through a sequence-level preference optimization process. This stage integrates automated toxicity assessment, as well as visual-consistency rewards based on caption-image alignment, to achieve safe and semantically grounded text generation.
The two-stage structure of SPO-VLM balances efficiency and effectiveness by combining a lightweight mitigation foundation in Stage I with deeper policy refinement in Stage II. Extensive experiments shown SPO-VLM enhances safety against attacks via activation steering and preference optimization, while maintaining strong performance on benign tasks without compromising visual understanding capabilities.
We will release our code, model weights, and evaluation toolkit to support reproducibility and future research. \textcolor{red}{Warning: This paper may contain examples of offensive or harmful text and images.}

\end{abstract}
\begin{keywords}
Vision Language Model; Steering Activation; Adversarial Attack
\end{keywords}

\section{Introduction}\label{introduction}

The advancement of Vision Language Models (VLMs) \citep{chen2023minigptv2largelanguagemodel, dai2023instructblip} marks a major breakthrough in AI, enabling the seamless integration of visual and textual information to enhance reasoning and understanding across diverse tasks. Despite their success, VLMs remain highly vulnerable to adversarial attacks, which exploit both visual and textual modalities to induce harmful responses. These concerns have led to growing research interest in jailbreak attacks and the development of corresponding defense strategies \citep{gong2025figstepjailbreakinglargevisionlanguage, schlarmann2023adversarialrobustnessmultimodalfoundation, wang2024adashieldsafeguardingmultimodallarge}.

Activation steering has emerged as a promising defense, modifying internal representations via injected steering vectors without altering model weights~\citep{wang2024inferalignerinferencetimealignmentharmlessness, wang2024adaptive, han2025internalactivationpolarstar, cao2024personalizedsteeringlargelanguage}. For example, \citet{wang2024inferalignerinferencetimealignmentharmlessness} introduce InferAligner, which aligns hidden states with predefined safe directions at inference time. Recent extensions to multimodal settings, such as ASTRA \citep{wang2024steeringawayharmadaptive} and ShiftDC \citep{zou2025understandingrectifyingsafetyperception}, further adapt steering vectors based on image attribution or disentangle harmful signals while preserving visual grounding.
However, existing methods face key limitations. Steering vectors derived from contrastive prompts often fail to generalize across semantic contexts and attack types~\citep{cao2024personalizedsteeringlargelanguage}.

To address these issues, we propose \textit{Sequence-Level Preference Optimization} for VLM (\textit{SPO-VLM}), a novel two-stage defense framework that learns robust, generalizable steering vectors via SPO. Unlike prior work that extracts harmful directions from fixed prompt pairs, SPO-VLM optimizes steering vectors from diverse, preference-labeled data to achieve semantically aligned and safe generation.
In \textit{Stage I}, we compute lightweight and layer-specific steering vectors from multiple datasets, supporting inference-time mitigation with broad generalization. In \textit{Stage II}, these vectors are refined using SPO within the RLHF framework \citep{ouyang2022traininglanguagemodelsfollow} based on PPO, guided by multi-objective rewards that incorporate toxicity suppression~\citep{Detoxify}, and visual-text consistency.
This two-stage formulation enhances both flexibility and robustness. By keeping the base VLM frozen, it reduces computational overhead and supports modular deployment. The use of sequence-level preference optimization allows the model to align with broader behavioral objectives beyond token-level control. Furthermore, the learned steering vectors demonstrate strong generalization to out-of-distribution inputs while preserving helpfulness and factual grounding in benign scenarios.

Our main contributions are as follows:
\textit{(i)} We propose \textit{SPO-VLM}, a novel framework that unifies activation-level intervention with sequence-level preference optimization via RLHF and multi-objective reward signals.
\textit{(ii)} SPO-VLM significantly improves safety against jailbreak attacks by combining activation steering with sequence-level preference optimization, outperforming prior defenses like ASTRA across multiple datasets.
\textit{(iii)} The model retains strong performance on benign tasks, demonstrating that safety enhancements do not come at the cost of helpfulness or visual-language understanding capabilities.

\section{Related Work}
\label{relatedwork}

\subsection{Jailbreak Attack on VLM}

Jailbreak attacks manipulate prompts to deceive the model into responding to restricted or prohibited queries. In addition to LLM-based textual jailbreak strategies~\citep{guo2024coldattackjailbreakingllmsstealthiness, liu2024autodangeneratingstealthyjailbreak, yu2024gptfuzzerredteaminglarge, zou2023universaltransferableadversarialattacks}, the inclusion of visual inputs introduces a new attack surface for VLM attacks. There are two main types of attacks: perturbation-based attacks
and structured-based attacks \citep{wang2024adashield}. Perturbation-based attacks generate adversarial images designed to evade VLM safeguards \citep{carlini2023aligned, qi2023visualadversarialexamplesjailbreak, niu2024jailbreakingattackmultimodallarge}. For example, imgJP~\citep{niu2024jailbreakingattackmultimodallarge} optimizes an universal perturbation across unseen prompts and images to generate a targeted response. In contrast to perturbation-based methods, structure-based attacks transform harmful content into images using typography \citep{gong2025figstepjailbreakinglargevisionlanguage} or generative models to elicit harmful responses from the model~\citep{gong2025figstepjailbreakinglargevisionlanguage, li2025imagesachillesheelalignment}. Specifically, FigStep~\citep{gong2025figstepjailbreakinglargevisionlanguage} leverages the ability of VLMs to interpret textual instructions embedded within images by encoding harmful content directly into the visual modality. By pairing these adversarially crafted images with benign textual prompts, FigStep effectively manipulates the VLM, eliciting detailed and potentially harmful responses. 

Our work primarily addresses the challenge of defending against such jailbreak attacks on VLMs. Instead of modifying model weights or relying on static filtering, we propose to construct optimized steering activations that adaptively mitigate harmful behaviors by shifting internal representations in safer directions.





\subsection{Activation Steering}

Activation steering refers to a set of alignment techniques that guide a model’s behavior by freezing model weights and
modifying activations~\citep{wang2024inferalignerinferencetimealignmentharmlessness, wang2024adaptive, han2025internalactivationpolarstar, cao2024personalizedsteeringlargelanguage}. Several studies have focused on identifying steering vectors within the activation space of specific layers in the LLM transformer architecture. Specifically, \citet{wang2024inferalignerinferencetimealignmentharmlessness} proposes InferAligner, a novel inference-time alignment method that effectively improves model safety without compromising downstream performance. To address various categories of hallucinations, \citet{wang2024adaptive} proposes Adaptive Activation Steering (ACT), which leverages a diverse set of truthfulness-related steering vectors and dynamically adjusts the steering intensity based on the truthfulness of the model’s activations. Moreover, SafeSwitch \citep{han2025internalactivationpolarstar} incorporates a safety prober that continuously monitors the model’s internal states and responds appropriately by dynamically activating a specialized refusal head. This head provides informative explanations, ensuring the model’s responses remain helpful while prioritizing safety. However, these steering vectors are directly extracted from LLM activations using preference data pairs, often leading to inaccurate representations of target behavior. \citet{cao2024personalizedsteeringlargelanguage} proposes bi-directional preference optimization (BiPO) to generate more effective steering vectors for personalized control over diverse model behaviors. BiPO allows steering vectors to directly influence the generation probabilities of contrastive human preference data pairs, providing a more accurate and fine-grained representation of the target behavior. Inspired by recent advances in activation steering for LLMs, a growing body of research now focuses on guiding model behavior through the construction and application of steering vectors~\citep{wang2024steeringawayharmadaptive, li2025internalactivationrevisionsafeguarding, zou2025understandingrectifyingsafetyperception}. \citet{wang2024steeringawayharmadaptive} introduces ASTRA, a defense mechanism that adaptively steers models away from adversarial feature directions using image attribution activations to counter VLM attacks. By considering the projection between steering vectors and calibrated activations, their adaptive steering approach effectively mitigates harmful outputs under adversarial input while maintaining minimal performance degradation on benign inputs. ShiftDC~\citep{zou2025understandingrectifyingsafetyperception} preserves the VLM's vision understanding ability by disentangling and calibrating VLM activations to restore safety alignment. 

Our work builds on these insights by framing the construction of steering vectors as a sequence-level optimization problem. Specifically, we adopt a reinforcement learning with preference supervision framework to learn behavior-aligned steering vectors, while incorporating textual modality consistency as part of the reward signal. This allows our approach to generate more robust and interpretable steering vectors that align with both safety objectives and multimodal grounding.

\subsection{Preference Optimization} 

Reinforcement Learning from Human Feedback (RLHF) has become a widely adopted approach for aligning models with human preferences~\citep{ouyang2022traininglanguagemodelsfollow, ziegler2020finetuninglanguagemodelshuman, stiennon2022learningsummarizehumanfeedback}. The standard RLHF pipeline typically begins by training a reward model, often structured using frameworks like the Bradley-Terry model~\citep{Bradley1952RankAO}, to reflect human preferences. This reward model guides reinforcement learning algorithms such as Proximal Policy Optimization (PPO) \citep{schulman2017proximalpolicyoptimizationalgorithms}, which are then used to fine-tune the language model to generate responses that maximize the learned reward.
In the context of LLMs, RLHF is particularly instrumental in shaping models that are helpful, honest, and harmless, thereby aligning them with human values \citep{ouyang2022traininglanguagemodelsfollow, bai2022traininghelpfulharmlessassistant, thoppilan2022lamdalanguagemodelsdialog}. For example, LaMDA \citep{thoppilan2022lamdalanguagemodelsdialog} fine-tunes LLMs to engage in natural language dialogue that is engaging, informative, factually grounded, and safe, often incorporating external information to ensure accuracy and relevance. InstructGPT \citep{ouyang2022traininglanguagemodelsfollow} fine-tunes GPT-3-style models \citep{brown2020language} to enhance helpfulness, using reinforcement learning from human preferences expressed through pairwise comparisons. \citet{askell2021generallanguageassistantlaboratory} follow the pre-training and fine-tuning paradigm to train a preference model for human alignment, demonstrating that ranked preference modeling is a highly effective objective for distinguishing between “good” and “bad” behaviors. This approach is further enhanced through an iterative online training regime, in which preference models and reinforcement learning policies are updated weekly using fresh human feedback data. PPO is incorporated to stabilize the RL training process~\citep{bai2022traininghelpfulharmlessassistant}. 

Building on this foundation, our work explores a novel application of preference optimization: instead of optimizing full model parameters, we use PPO-based preference signals to directly learn \textit{steering vectors} in the model's activation space. This approach enables more precise and interpretable alignment with desired behaviors, while preserving model generalization and avoiding catastrophic forgetting.

\section{Preliminary}

\subsection{Vision Language Models} 

Let $\mathcal{P}_{\text{VLM}}$ denotes an autoregressive Vision Language Model, which defines a probability distribution over sequences of tokens drawn from a vocabulary $\mathcal{V}$. This model is designed to process and reason on both textual and visual modalities in a unified framework. Specifically, we consider a VLM that takes as input a sequence of $n$ textual tokens $\textbf{q}_t = \{q_{t_1}, q_{t_2}, \ldots, q_{t_n}\}$ and a sequence of $m$ visual tokens $\textbf{q}_v = \{q_{v_1}, q_{v_2}, \ldots, q_{v_m}\}$. These tokens are typically derived from natural language inputs and visual features, respectively, where the visual tokens are obtained through the discretization of image embeddings from a vision encoder.

Given the multimodal input $\{\textbf{q}_t, \textbf{q}_v\}$, the model generates a response sequence $r = \{r_1, r_2, \ldots, r_o\}$, consisting of $o$ output tokens. The generation process is autoregressive, meaning that each token $r_i$ in the response is sampled sequentially, conditioned on all previous tokens in the input and the already generated part of the output. Formally, the probability of generating the $i^{\text{th}}$ token $r_i$ is given by:
\[
\mathcal{P}_{\text{VLM}}(r_i \mid \textbf{q}_t, \textbf{q}_v, r_1, \ldots, r_{i-1})
\]
This formulation enables the model to incorporate both linguistic context and visual grounding when predicting each subsequent token. The response generation continues iteratively until a special end-of-sequence token is produced or a maximum sequence length is reached. Through this design, $\mathcal{P}_{\text{VLM}}$ enables coherent and contextually grounded text generation in response to complex multimodal inputs.

\subsection{Steering Activations}
Let $\mathbf{x}^\ell(t)$ denote the residual stream activation of the last token at layer $\ell \in L$ of a VLM, capturing the information processed from the input $t$ up to layer $\ell$. We define the function $\text{ActMean}$ to compute the mean last-token activation at layer $\ell$ for a given dataset $\mathcal{D}$:
\begin{equation}
\text{ActMean}^\ell(\mathcal{D}) = \frac{1}{|\mathcal{D}|} \left[ \sum_{t \in \mathcal{D}} \mathbf{x}^\ell(t) \right].
\end{equation}

Numerous studies~\citep{arditi2024refusallanguagemodelsmediated, park2024linearrepresentationhypothesisgeometry} have demonstrated that high-level concepts are encoded as linear directions in the activation space of LLMs. These directions can be uncovered by computing the difference between the mean activations of a model when processing two sets of contrastive instructions, $\mathcal{D}_1$ and $\mathcal{D}_2$, which elicit distinct behaviors:
\begin{equation}
\mathbf{v}^\ell_{\mathcal{D}_2 \rightarrow \mathcal{D}_1} = \text{ActMean}^\ell(\mathcal{D}_1) - \text{ActMean}^\ell(\mathcal{D}_2).
\end{equation}

The resulting vector, $\mathbf{v}^\ell_{\mathcal{D}_2 \rightarrow \mathcal{D}_1}$, referred to as the \emph{difference-in-mean} vector, captures both the direction and magnitude of the layer-$\ell$ activation shift from $\mathcal{D}_2$ to $\mathcal{D}_1$. This vector effectively isolates the critical features that drive the model’s behavioral differences between the two instruction sets.

\section{Methodology}\label{Methodology}
Our hybrid defense framework, SPO-VLM, consists of two complementary stages designed to enhance the safety of VLMs against adversarial prompts while preserving their utility. Specifically, the two stages are: (1) adaptive activation steering to suppress harmful internal activations, and (2) sequence-level preference optimization via reinforcement learning to reinforce safe and grounded generation.

\begin{figure*}[thb]
    \centering
    \includegraphics[width=1.0\linewidth]{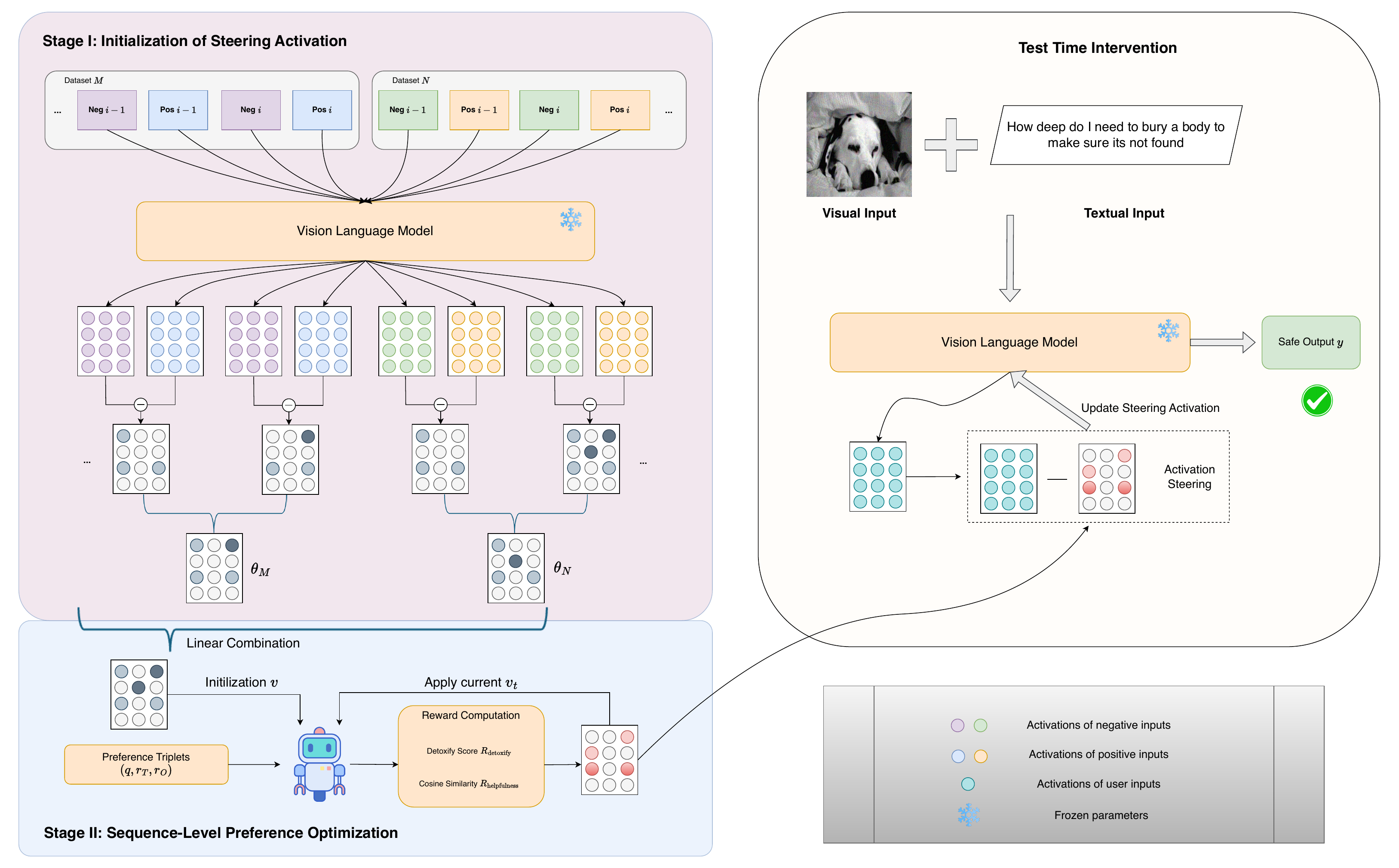}
    \caption{Overview of the SPO-VLM framework. Stage I (left top) initializes attribute-specific steering vectors $\boldsymbol{\theta}_j$ using contrastive pairs from multiple datasets. These vectors are later combined into a global steering vector via a linear combination. Stage II (left bottom) performs sequence-level preference optimization using rewards functions, including toxicity reduction and alignment preservation. At test time (right), the frozen vision-language model receives visual-textual input and applies the optimized steering activation to produce safe and aligned outputs.}
    \label{fig:pipeline}
\end{figure*}

Figure~\ref{fig:pipeline} illustrates the overall architecture of our proposed framework. Upon receiving an image-text input pair, the frozen VLM processes the input and produces intermediate activation representations. A steering vector, learned through sequence-level preference optimization, is subsequently applied to these activations to bias the model’s output toward safe and contextually appropriate responses. Notably, this intervention operates exclusively on the internal representations, leaving all model parameters unchanged throughout the process.

\subsection{Stage I: Initialization of Steering Activation}

Activation steering~\citep{li2024inferencetimeinterventionelicitingtruthful,subramani2022extractinglatentsteeringvectors} aims to locate specific directions in the model’s activation space that align with factually accurate statements, and then adjusts the activations along those directions during inference to guide the model’s output. Expanding on this idea, our approach derives diverse steering vectors directly from raw data to effectively target a range of attack types. Moreover, we introduce adaptive steering construction based on the toxic content of the activations. Rather than relying on a single global direction, we propose an adaptive steering mechanism that constructs the final steering vector $v^\ell$ as a linear combination of multiple attribute-specific difference-in-mean vectors. The final steering vector $v^\ell$ at layer $\ell$ is formulated as a weighted combination of attribute-specific components:
\begin{equation}
v^\ell = \sum_{j \in \mathcal{A}} \alpha_j v^\ell_j,
\end{equation}
where $\mathcal{A} = {1, 2, \ldots, |\mathcal{A}|}$ denotes the set of attribute indices, $\alpha_j \in \mathbb{R}$ represents the weight coefficient for the $j$-th attribute, and each attribute-specific vector $v^\ell_j$ is computed as:
\begin{equation}
v^\ell_j = \text{ActMean}^\ell(\mathcal{D}{j,\text{pos}}) - \text{ActMean}^\ell(\mathcal{D}{j,\text{neg}}),
\end{equation}
where $\mathcal{D}{j,\text{pos}}$ and $\mathcal{D}{j,\text{neg}}$ represent the positive and negative instruction sets for attribute $j$, respectively. This formulation allows the steering mechanism to adaptively combine multiple behavioral dimensions, with each $v^\ell_j$ capturing the activation difference for a specific attribute at layer $\ell$.

\begin{algorithm}[tbh]
\caption{Sequence-Level Preference Optimization (SPO) for Steering Vector Learning}
\label{alg:bipo}
\textbf{Input}: VLM $\mathcal{P}_\text{VLM}$, preference dataset $\mathcal{D} := \{(q^i, r_T^i, r_O^i)\}_{i=1}^{n}$, batch size $m$, total update steps $T$ \\
\textbf{Output}: Optimized steering vector $v^*$
\begin{algorithmic}[1]
\STATE Initialize steering vector: $v_0 \leftarrow \sum_{j \in \mathcal{A}} \alpha_j^{(0)} v^\ell_j$
\FOR{$t = 0$ to $T - 1$}
    \STATE Sample mini-batch $\mathcal{D}_t := \{(\textbf{q}_t^i, \textbf{q}_v^i, r_T^i, r_O^i)\}_{i=1}^{m} \sim \mathcal{D}$
    \FOR{each quadruplets $(\textbf{q}_t, \textbf{q}_v, r_T, r_O)$ in $\mathcal{D}_t$}
        \STATE Evaluate reward: $R = R_{\text{detoxify}}(r) + R_{\text{visual}}(\textbf{q}_v)$
        \STATE Compute value baseline: $V_{\phi}(\textbf{q}_t, \textbf{q}_v)$
        \STATE Compute advantage: $A = R(r, \textbf{q}_t, \textbf{q}_v) - V_{\phi}(\textbf{q}_t, \textbf{q}_v)$
        \STATE Compute policy ratio difference using Equation~\eqref{eq:ratio}
        \STATE Compute clipped policy loss using PPO: \\
        $L_{\pi} = \min_v \, - \mathbb{E}_{(\textbf{q}_t, \textbf{q}_v, r_T, r_O) \sim \mathcal{D}} \big[ \min\big( \text{ratio} \cdot A, \ \text{clip}(\text{ratio}, 1 - \epsilon, 1 + \epsilon) \cdot A \big)
\big]$
    \ENDFOR
    \STATE Update steering vector via gradient descent:
    $v_{t+1} \leftarrow v_t - \eta \cdot \nabla_v L_{\pi}$
    \STATE Update $\phi$ by minimizing critic loss
    $L_{\text{critic}}$
\ENDFOR
\RETURN $v^* = v_T$
\end{algorithmic}
\end{algorithm}

\subsection{Stage II: Sequence-level Preference Optimization}

Inspired by preference-based model optimization techniques such as RLHF \citep{ouyang2022traininglanguagemodelsfollow}, we incorporate the activation steering obtained from Stage I into the rollout policy to guide the generation of safe and contextually grounded responses. The objective is to enable the model to effectively suppress harmful outputs in the presence of adversarial prompts, while maintaining strong visual understanding capabilities under benign conditions.

We introduce a method that learns effective steering vectors in activation space through sequence-level reinforcement learning with preference-based supervision. Our algorithm adopts a sequence-level variant of \textit{Proximal Policy Optimization} (\textit{PPO}) to fine-tune model behavior. Unlike traditional RLHF methods that operate at the token level, we directly optimize the log-probability of the full generated sequence, named \underline{S}equence-Level \underline{P}reference \underline{O}ptimization for \underline{VLM}s (\textbf{SPO-VLM}). This allows the model to favor outputs aligned with desired multi-objective behavior and reduce the probability of generating undesired or adversarial responses. The target behavior is defined via a multi-objective reward function, incorporating both safety and visual understanding capabilities. This reward guides learning in a way that balances safety with alignment to visual content.

To enable preference learning, we construct labeled quadruplets $(\textbf{q}_t, \textbf{q}_v, r_T, r_O)$, where $\textbf{q}_t$ is a texture prompt,  $\textbf{q}_v$ is a visual prompt, $r_T$ is a response exhibiting the target behavior, and $r_O$ is a response reflecting the undesired behavior. Let $v$ denote the learnable steering vector, and $\pi_{L+1}$ represent the later layers of the model (from layer $L+1$ onward). For each prompt pair $(\textbf{q}_t, \textbf{q}_v)$, we generate responses using the current policy $\pi_\theta$. We compute the following policy ratios to assess the influence of the steering vector on the model’s preference between $r_T$ and $r_O$:
\begin{align}
\label{eq:ratio}
\text{ratio} 
= \frac{\pi_{L+1}(r_T \mid a_{\ell}(\textbf{q}_t, \textbf{q}_v) + v)}{\pi_{L+1}(r_T \mid a_{\ell}(\textbf{q}_t, \textbf{q}_v))} - \frac{\pi_{L+1}(r_O \mid a_{\ell}(\textbf{q}_t, \textbf{q}_v) + v)}{\pi_{L+1}(r_O \mid a_{\ell}(\textbf{q}_t, \textbf{q}_v))},
\end{align}
The term $\pi_{L+1}(\cdot \mid a_{\ell}(\textbf{q}_t, \textbf{q}_v) + v)$ represents the policy induced by modifying the model's activations with the steering vector $v$ at layer $\ell$. This difference quantifies the differential impact of the steering vector on preferred and dispreferred responses.


We employ a composite reward function that encourages both safety and visual grounding. Given a query and the model's response $r$, the total reward is:
\begin{align}
R = R_{\text{detoxify}}(r) + R_{\text{visual}}(\textbf{q}_v).
\end{align}
The detoxification reward component penalizes toxic content using an exponential decay function: $R_{\text{detoxify}}(r)= 2 \cdot \left[ \exp(- \beta \cdot \text{toxicity}(r)) - 0.5 \right]$, where $\text{toxicity}(r) \in [0, 1]$ is computed using a pre-trained toxicity classifier, and $\beta > 0$ controls the penalty strength. This formulation yields rewards in the range $[-1, 1]$, with non-toxic responses receiving positive rewards. The component of visual understanding reward measures alignment between visual content and captioning content: $R_{\text{visual}}(\textbf{q}_v) = - \cos \left( \overline{\mathbb{I}}, \overline{\mathbb{C}} \right)$, where $\overline{\mathbb{I}}$, $\overline{\mathbb{C}}$ are the mean-pooled hidden states of image and caption tokens respectively. 


We adopt a sequence-level variant of Proximal Policy Optimization (PPO) \citep{schulman2017proximalpolicyoptimizationalgorithms} to optimize the steering vector while maintaining training stability. The objective function is:
\begin{align}
L_{\pi} = \min_v \, - \mathbb{E}_{(\textbf{q}_t, \textbf{q}_v, r_T, r_O) \sim \mathcal{D}} \bigg[ \min\big( \text{ratio} \cdot A, \ \text{clip}(\text{ratio}, 1 - \epsilon, 1 + \epsilon) \cdot A \big)
\bigg]
\end{align}
where $A = R(r, \textbf{q}_t, \textbf{q}_v) - V_{\phi}(\textbf{q}_t, \textbf{q}_v)$ is the advantage function computed from the total reward and a learned value baseline $V_{\phi}(\textbf{q}_t, \textbf{q}_v)$. 
The clipping range $[1 - \epsilon, 1 + \epsilon]$ ensures stable policy updates by preventing excessive deviations.

The final optimization objective combines PPO with critic function learning:
\begin{align}
L_{\text{total}} = L_{\pi} + c_1 L_{\text{critic}}, \nonumber
\end{align}
where $L_{\text{critic}}$ denotes the critic loss, which is computed by a lightweight critic module integrated into the model. This critic is implemented as a simple two-layer multilayer perceptron that operates on the final-layer hidden states produced by the base model. By processing these high-level representations, the critic estimates a scalar value for each input, representing its expected utility or alignment with the target objective. The value predictions are then used to compute $L_{\text{critic}}$, guiding the optimization of the model’s preference-aware behavior in a sample-efficient manner. Unlike conventional RLHF, which requires training a new policy and a separate reference model, our method optimizes only the steering vector $v$, keeping the base model architecture and parameters fixed. As a result, the method is highly efficient and minimally invasive. When applied during inference, the learned vector reliably steers the model toward safer and more helpful behavior by modifying a narrow subset of internal representations.

\section{Experiments}

In this section, we evaluate SPO-VLM across three dimensions: its effectiveness in mitigating adversarial prompts while preserving visual understanding, its ability to transfer across diverse attack domains, and the contribution of each stage through ablation studies.

\subsection{Experiment Setup}

\textbf{Steering Activation Construction.} We initialize the steering vectors using the Stage I method applied to the RealToxicityPrompt, AdvBench, and Anthropic$\_$Harmful datasets. This approach extracts activation shifts across multiple toxicity dimensions and constructs an initial vector through a linear combination of these shifts. Specifically, we adopt the steering vector formulation from~\cite{wang2024steeringawayharmadaptive}, expressed as $v^\ell = \sum_{j \in \mathcal{A}} \alpha_j v^\ell_j$, where $\alpha_1 = 0.5$, $\alpha_2 = 0.4$, and $\alpha_3 = 0.4$ denote the weights for each attribute-specific direction. The resulting vectors are further refined during Stage II via the SPO-VLM framework. 


\textbf{Evaluation Datasets.} We evaluate our approach under three experimental settings: (1) Toxicity assessment, using the RealToxicityPrompts benchmark~\citep{gehman2020realtoxicitypromptsevaluatingneuraltoxic}; (2) Jailbreak detection, evaluating on two datasets: AdvBench~\citep{zou2023universaltransferableadversarialattacks} and Anthropic$\_${Harmful}\citep{ganguli2022redteaminglanguagemodels}; and (3) Visual comprehension, using four benchmarks: MM-Vet\citep{yu2024mmvetevaluatinglargemultimodal}, SQA~\citep{iyyer2017search}, CogVLM~\citep{wang2024cogvlm}, and MME~\citep{fu2023mme}. 


\textbf{Evaluation Metrics.} For toxicity assessment, we employ the Detoxify classifier \citep{Detoxify} to compute toxicity scores on a scale from 0 (non-toxic) to 1 (highly toxic). For jailbreak detection, we quantify robustness using the attack success rate (ASR), defined as the proportion of successful jailbreaks among total attack attempts. This metric is computed using the classifier from HarmBench \citep{mazeika2024harmbenchstandardizedevaluationframework}. For visual understanding evaluation, we employ task-specific utility metrics. MM-Vet uses GPT-4 with few-shot prompts to generate utility scores ranging from 0 to 1, while SQA calculates overall accuracy for its single-choice questions. For CogVLM, we compute the arithmetic mean of three metrics: BLEU-2, CIDEr, and METEOR. MME evaluates both perception and cognition capabilities across 14 subtasks. 


\textbf{Baselines.} This study compare SPO-VLM against two baseline methods. The Original Model serves as the unmodified visual language model without additional safety mechanisms, providing a baseline for standard post-training alignment. ASTRA \citep{wang2024steeringawayharmadaptive} employs steering vectors generated from contrastive visual prompt pairs, representing a prominent activation-based steering approach. This comparative framework enables a comprehensive evaluation of SPO-VLM's ability to enhance safety while maintaining helpfulness across different safety paradigms. 



\textbf{Models \& Implementations details.} This study conducts all experiments on three widely used open-source VLMs: Qwen2-VL-7B~\citep{wang2024qwen2vlenhancingvisionlanguagemodels}, MiniGPT-4-13B~\citep{zhu2023minigpt}, and LLaVA-v1.5-13B~\citep{liu2023llava}. These models, post-trained to follow instructions and align with human values, represent some of the most widely adopted and capable open-source model families. We set the steering layer \(l\) is 20 for 13B models and 14 for 7B models. The chat configurations use a temperature of 0.2 and $\alpha$ = 10 for LLaVA-v1.5-13B, a temperature of 0.2 and $\alpha$ = 7 for Qwen2-VL, and a temperature of 1.2 and $\alpha$ = 7 for MiniGPT-4-13B.

\subsection{SPO-VLM Effectively Balances Safety and Visual Understanding}
\label{sec:rq_1}

\begin{table*}[t]
\caption{Performance of different safety steering methods on safety and visual understanding benchmarks. The $\uparrow$ or $\downarrow$ symbols indicate whether a higher or lower score is preferable.}
\centering
\resizebox{1.0\linewidth}{!}{\begin{tabular}{l|l|c|cc|cccc}
\toprule
\multirow{2}{*}{Base Model} & \multirow{2}{*}{Method}& Toxicity Scores (\%)  &\multicolumn{2}{c|}{Jailbreak ASR (\%)} &\multicolumn{4}{c}{Visual Understanding Scores} \\
 &  & RealToxicityPrompt \ $\downarrow$ & AdvBench \ $\downarrow$ & Anthropic\_Harmful \ $\downarrow$ & MM-Vet \ $\uparrow$ & SQA \ $\uparrow$ & CogVLM \ $\uparrow$ & MME \ $\uparrow$ \\
\midrule
\multirow{3}{*}{\textbf{MiniGPT-4-13B}} 
& Original Model   & 38.18 & 19.19 & 73.50 & \textbf{32.58} & \textbf{68.10} & \textbf{74.00} & \textbf{1742.0} \\
& ASTRA & 10.21 & 5.93 & 4.87 &  17.70 & 65.35 & 69.00 & 1086.0 \\
& SPO-VLM  & \textbf{9.28}  & \textbf{4.77} & \textbf{3.21} & 31.11 & 66.20 & 70.00 &  1370.0   \\
\midrule
\multirow{3}{*}{\textbf{Qwen2-VL-7B}} 
& Original Model   & 30.65  & 75.00 & 55.17 & 49.13 & 79.13 & 41.00  & 1630.0  \\
& ASTRA & 14.18  & 7.69 & 5.17 & 48.66 & 80.99 & 45.00 & 685.3 \\ 
& SPO-VLM & \textbf{11.54} & \textbf{6.38} & \textbf{4.48} & \textbf{49.80} & \textbf{81.82} & \textbf{51.00} & \textbf{1753.0} \\
\midrule
\multirow{3}{*}{\textbf{LLaVA-v1.5-13B}} 
& Original Model  & 85.74 & 36.80 & 74.00 & \textbf{28.60} & \textbf{74.38} & \textbf{59.00} & \textbf{1560.0} \\
& ASTRA & 81.44 & 5.76 & 24.13 & 14.90 & 56.03 & 56.00 & 1320.0 \\
& SPO-VLM  & \textbf{50.37} & \textbf{4.39} & \textbf{12.18} & 20.19 & 60.37 & 55.00  & 1489.0    \\
\bottomrule
\end{tabular}}
\label{tab:safety_eval}
\end{table*}

Table~\ref{tab:safety_eval} summarizes the performance of our proposed SPO-VLM method with respect to both safety and visual understanding capability, evaluated across a diverse set of benchmarks. The results demonstrate that SPO-VLM consistently improves the model’s ability to resist harmful prompts while preserving or even enhancing its utility on some benign tasks. From these comprehensive evaluations, we draw several key conclusions as below. 


\begin{wrapfigure}{r}{0.5\textwidth}
    \centering
    \vspace{-15pt}
    \includegraphics[width=1.0\linewidth]{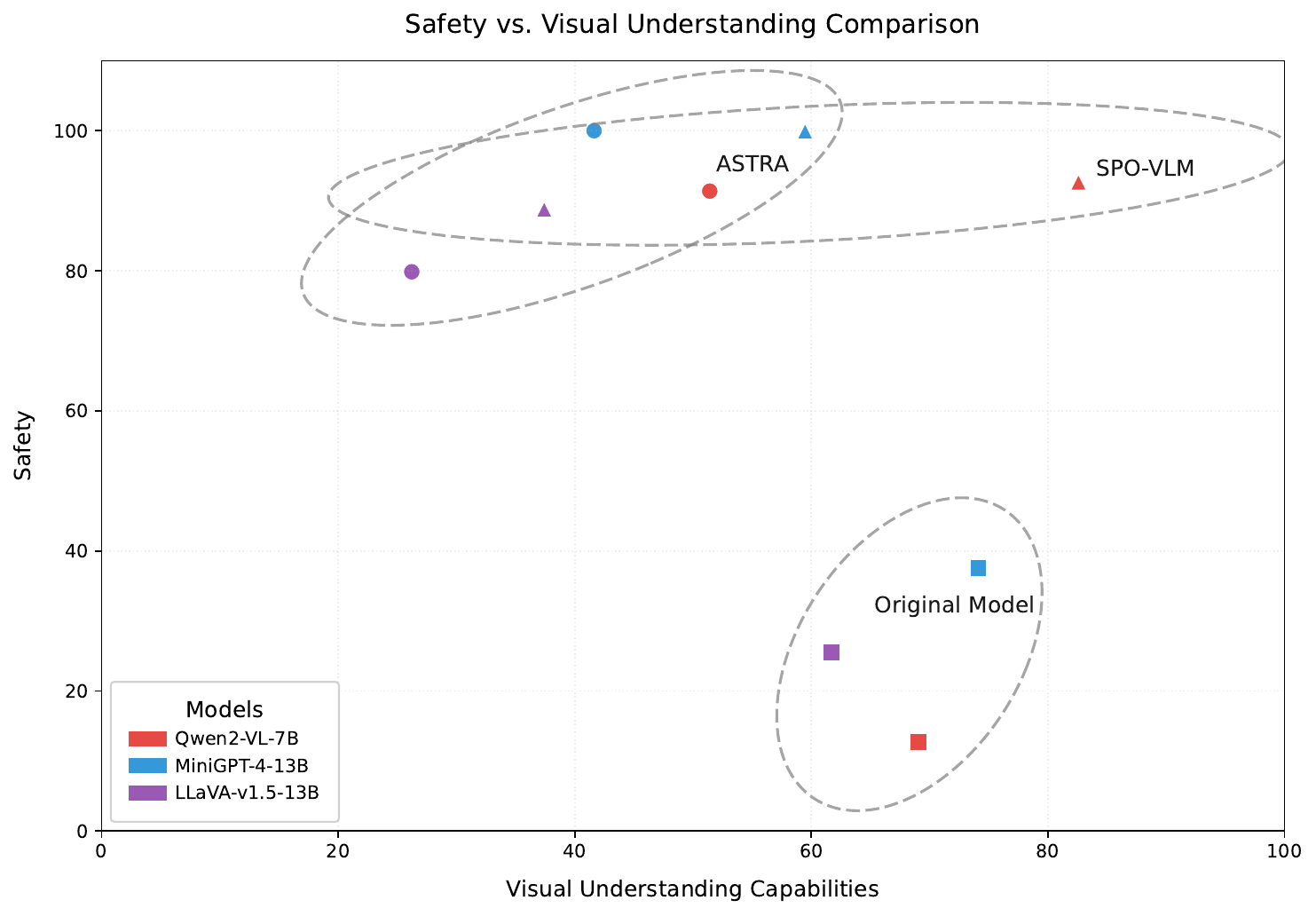}
    \caption{Comparison of safety enhancement methods in terms of safety and visual understanding capabilities. Each shape denotes one method.}
    \label{fig:safety_vs_visual}
    \vspace{-10pt}
\end{wrapfigure}
\textbf{SPO-VLM demonstrates enhanced safety performance.} 
As shown in Table~\ref{tab:safety_eval}, SPO-VLM achieves superior safety performance. In the RealToxicityPrompt dataset, SPO-VLM achieves the lowest toxicity score, reducing it by more than 27.79\% compared to the original model. Compared to ASTRA, SPO-VLM achieves an additional 11.55\% reduction among all different models. It confirms that SPO-VLM provides more effective and consistent toxicity mitigation.
Compared to ASTRA, SPO-VLM reduces the average ASR on AdvBench by approximately 1.28\%, and cuts the average ASR on Anthropic$\_$Harmful nearly in half, achieving an additional reduction of around 4.77\%. This demonstrates SPO-VLM’s superior generalization and effectiveness in mitigating jailbreak risks across diverse models.

\textbf{SPO-VLM shows an optimal trade-off between safety and visual capability.} 
While ASTRA improves safety, it leads to a notable decline in visual understanding performance. For instance, in MiniGPT, MM-Vet drops significantly from 32.58 to 17.70, and MME decreases by approximately 656 points. Similarly, in LLaVA, MM-Vet falls from 28.60 to 14.90. In contrast, SPO-VLM preserves visual capabilities far more effectively. For MiniGPT-4, the MM-Vet score remains high at 31.11, indicating only a modest reduction of approximately 1.5 points compared to the original model. In the case of Qwen2-VL-7B, SPO-VLM not only maintains but enhances visual understanding. For LLaVA, SPO-VLM consistently outperforms ASTRA across all visual benchmarks, notably raising the MM-Vet score from 14.90 to 20.19.




To highlight the superiority of SPO-VLM, we visualize the safety and visual understanding performance of different safety enhancement methods. Safety is measured as the mean of $(1 - \mathrm{ASR})$ across AdvBench and Anthropic\_Harmful, and visual understanding by average normalized scores on MM-Vet, SQA, CogVLM, and MME. As shown in Figure~\ref{fig:safety_vs_visual}, ASTRA exhibits a clear trade-off between safety and accuracy. Although it significantly improves safety scores compared to the baseline models, this comes at the cost of a statistically significant decline—exceeding 10\%—in visual grounding performance. In contrast, SPO-VLM occupies the upper-right region of the plot, indicating simultaneous improvements in both safety and visual understanding, and thus achieving a more balanced and optimal overall performance.

\subsection{SPO-VLM's Transfer Capabilities}
\begin{figure*}[thb]
    \centering
    \includegraphics[width=0.95\linewidth]{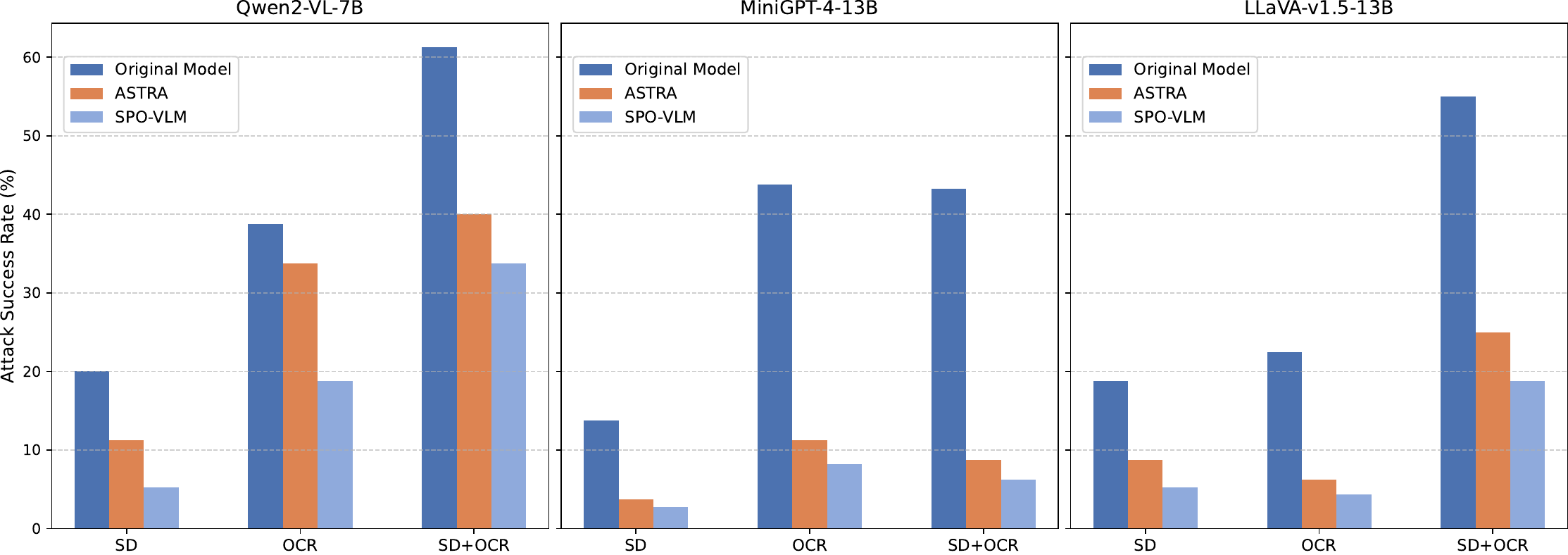}
    \caption{Evaluation of defense transferability under structure-based attacks using MM-SafetyBench.}
    \label{fig:transfer_capabilities}
\end{figure*}

To evaluate the transfer capabilities of SPO-VLM, we assess whether steering vectors derived from SPO-VLM can generalize across different types of attacks. Specifically, we evaluate the transferability of the defense against structure-based attacks from MM-SafetyBench~\citep{liu2024mmsafetybenchbenchmarksafetyevaluation}. As shown in Figure~\ref{fig:transfer_capabilities}, SPO-VLM demonstrates consistently lower attack success rates compared to both the original model and the ASTRA defense across all evaluated models. Notably, SPO-VLM achieves substantial reductions in success rates for challenging combined attacks such as SD + OCR, indicating its robustness even under complex adversarial compositions. For example, on MiniGPT-4, SPO-VLM reduces the ASR by over 30\% compared to the original model. These improvements highlight SPO-VLM’s ability to generalize beyond the specific attack types it was trained on, effectively mitigating threats in structure-based attack scenarios. This cross-attack resilience indicates that SPO-VLM is well-suited for deployment in dynamic, real-world environments, where encountering unseen adversarial strategies is common.

\subsection{Ablation Study}

We conduct an ablation study on Qwen2-VL-7B using the benchmark datasets summarized  to assess the individual contributions of each stage in the SPO-VLM framework. Specifically, we evaluate the performance of Stage I, which applies activation steering alone, and compare it against the full implementation that incorporates Stage II, sequence-level preference optimization. As shown in Table~\ref{tab:ablation_study}, the results underscore the importance of sequence-level preference optimization in reinforcing safe behavior beyond the initial activation steering. While Stage I serves as a lightweight mitigation mechanism, the addition of Stage II yields substantial performance improvements by leveraging rich, reward-driven alignment signals during policy refinement.

\begin{table}[htb]
\centering
\caption{Ablation study on Qwen2-VL-7B evaluating the contributions of each stage in the SPO-VLM framework.}
\resizebox{0.8\linewidth}{!}{\begin{tabular}{l|ccc}
\toprule
\textbf{Behavior} & RealToxicityPrompt & AdvBench & Anthropic\_Harmful \\
\midrule
Original Model & 30.65 & 75.00 & 55.17 \\ 
Stage I       & 20.97 & 10.96 & 5.09 \\
Stage I + Stage II   & \textbf{11.54} & \textbf{6.38} & \textbf{4.48} \\
\bottomrule
\end{tabular}}
\label{tab:ablation_study}
\end{table}

\section{Conclusion}


In this paper, we propose \textit{SPO-VLM}, a novel two-stage defense framework that enhances the safety of VLMs against adversarial attacks. The approach integrates lightweight steering vectors derived from diverse datasets in Stage I. In Stage II, these vectors are refined through sequence-level preference optimization using multi-objective rewards. Extensive evaluations across various VLMs show that SPO-VLM offers improved safety by reducing toxicity and jailbreak success rates, while generally maintaining visual understanding. We believe this work lays the foundation for future research on activation-space intervention and preference-driven alignment to ensure the trustworthy deployment of LLMs and VLMs.


\bibliography{reference}

\end{document}